# Mastering Large Scale Multi-label Image Recognition with high efficiency over Camera trap images.

## Hakuna Ma-data Challenge: 1st place submission description

**Miroslav Valan**
*Swedish Museum of Natural History*
*Stockholm University*
*Savantic AB*
*PiVa AI*
miroslav.valan@nrm.se

**Lukáš Picek**
*University of West Bohemia*
*PiVa AI*
picekl@kky.zcu.cz

## Abstract

*Camera traps are crucial in biodiversity motivated studies, however dealing with large number of images while annotating these data sets is a tedious and time consuming task. To speed up this process, Machine Learning approaches are a reasonable asset. In this article we are proposing an easy, accessible, light-weight, fast and efficient approach based on our winning submission to the "Hakuna Ma-data - Serengeti Wildlife Identification challenge". Our system achieved an Accuracy of **97%** and outperformed the human level performance. We show that, given relatively large data sets, it is effective to look at each image only once with little or no augmentation. By utilizing such a simple, yet effective baseline we were able to avoid over-fitting without extensive regularization techniques and to train a top scoring system on a very limited hardware featuring single GPU (1080Ti) despite the large training set (6.7M images and 6TB).*

## 1. Introduction

Camera traps are an immense resource in ecological research and conservation efforts in addressing the need for accurate assessments of wildlife populations. They are used for passively collecting animal behavior data with minimal if any disturbance. Such device is activated remotely with a sensor - based on motion or an infrared sensor - or a beam trigger based on light, laser or sound. In addition to animals, sensors are often falsely triggered, for instance by moving grass, thus increasing the cost of processing already large number of images several times (4x in case of the Snapshot Serengeti) [1].

For these reasons, a system that can automatically analyze camera trap images, while maintaining the accuracy of humans, would save years of human effort and unlock new research opportunities in biodiversity. Recent work [2, 3] showed that modern Convolutional Neural Networks (CNNs) are achieving very promising results and closing the gap with the human level performance (3.5% error rate). [4]. Furthermore, when such CNN is utilized within human-in-the-loop approach the error rate might be reduced to 0.22%.

The main contribution of this paper is a simple yet effective training method, that learns representations from camera trap images by looking at every image only once (**1 epoch only**) while using no augmentations. Our proposed system won **1st place** on Hakuna Ma-data Challenge. Unlike previous work [2, 3] on the same Snapshot Serengeti data set [1] where empty images are 1) filtered and if animal is present then 2) classified as one of the animals, our system uses a single step, while being able to **predict multiple species** on the same image. In addition, our approach provides significant performance boosts being the **first ever to outperform humans** on this task with low complexity models suitable for **embedded edge devices**.

## 2. Hakuna Ma-data Challenge

Snapshot Serengeti is the largest camera trap project to date running continuously in Serengeti National Park, Tanzania, for more than 45 years [1]. The project is currently using a network of 225 cameras with passive infrared sensors that trigger every time a warm object moves in front of them. Each such event has one to several images (average 3) and is provided as a sequence to annotators. A total, 7.1M images from 2.65M sequences is currently annotated with an estimated effort of close to 20 man-years of work[1].

The Hakuna Ma-data challenge was designed to test cur-rent state-of-the-art approaches over the Snapshot Serengeti data set. The task was to predict presence of animals and categorize them into one of the 54 categories. The categories could be of different granularity; mostly at the species level, but could be higher taxonomic ranks (reptiles, other birds) or from the same species (lion female and lion male).

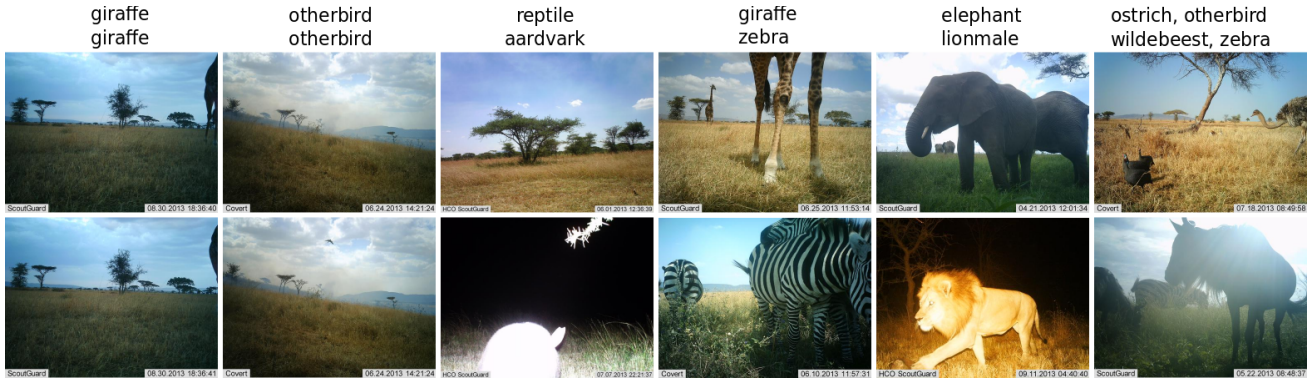

Figure 1. Example images from Snapshot Serengeti data set. The examples show a range of difficulties from severe occlusions on the left, through tiny objects (small size or object far from camera), close-up shots, motion blur and multiple classes on the same image.

Organizer provided 6.7M for training while keeping approximately 0.5M for testing. Competitors had no access to test images nor their annotations and needed to submit a Dockerized solution that run on the server.

The training set is largely imbalanced with 3/4 of images being empty with no animal present on them. Presence of different categories in the training set was also hugely imbalanced toward wildebeests, zebras and Thomson's gazelles.

## 3. Methodology

The proposed system is based on 3 architectures that were fine-tuned from the publicly available checkpoints pretrained on ImageNet [5]. SE-ResNext50 [6][7][8], EfficientNet B1 and EfficientNet B3 [9]. All the networks shared the Parameters from Table 1 and used one cycle training policy [10, 11, 12].

| Parameter | Value |
|---|---|
| Optimizer | Adam[13] |
| Warm start Learning rate | 600 iterations * BS |
| Maximum learning rate | 0.0001 |
| End learning rate | 0.000001 |
| Learning rate decay type | cosine annealing |

Table 1. Optimizer hyper-parameters identical to all networks.

Other parameters such us Batch size, Gradient Accumulation and Sampling strategy were set differently for each model as listed in Table 2.

| # | Architecture | BS | GA | Sampling strategy |
|---|---|---|---|---|
| 1 | SE-ResNext50 | 16 | 2 | season_by_season |
| 2 | SE-ResNext50 | 13 | 3 | random |
| 3 | EfficientNet B1 | 16 | 2 | season_by_season |
| 4 | EfficientNet B3 | 11 | 3 | random |

Table 2. Networks and hyper-parameters used in the experiments. BS = Batch Size. GA = Gradient Accumulation.

Specifically, models were trained for single epoch only using two sampling strategies. In one strategy we used random sampling from all seasons and in the other we processed data in chunks (subsets) composed of one or more seasons. The seasons 9 and 10 were purposely used at the end of the training to allow the model to get a better understanding of the recent changes in the vegetation.

For the final submission we used an ensemble on two SE-ResNext50 and single B1 and B3 models while combining their predictions with mean, geometrical mean and category aware average.

We use sigmoid function to enable detection of multiple animal categories from the same image. For EfficientNets we kept the default Dropout [14] of 0.2 and 0.3 for B1 and B3 respectively and for SE-ResNext-50 we set it to 0.0.

Due to the circular nature of the data (both time and date), and varying weather conditions (clouds, shades, wind) we might avoid most of the augmentations. We used only the random horizontal flips.

## 4. Results

Our final submission was a weighted ensemble of 4 models with total of 6 forward passes per test image. For each image we calculated the arithmetic mean for every animal category and geometric mean for the empty category which we refer as "class aware average" in Table 3. For sequences with 2 and more images we obtained the final predictions by calculated the arithmetic mean.

| # | Team Name | Private Score |
|---|---|---|
| 1 | **(ours)** | **0.00531** |
| 2 | n01z3 | 0.00532 |
| 3 | bragin | 0.0056 |
| 4 | green_elephant | 0.0055 |
| 5 | Appsilon | 0.0068 |
| 6 | BGU_DS | 0.0077 |
| 7 | saket | 0.0085 |
| 8 | MichalBusta | 0.0092 |
| 9 | bonlime | 0.0103 |
| 10 | DeepZebra | 0.0111 |
| - | *Baseline* | 0.0469 |

Table 3. Results of the top ten teams in Hakuna Ma-data challenge calculated as mean aggregated logarithmic loss (AggLogLoss).

Our submission to the challenge achieved the best scores in terms of Aggregated Logaritmical Loss (AggLogLoss). The AggLogLoss is defined as follows: For each possible category in a sequence, the binary log loss will be computed and then the results will be summed. The sum of the binary losses represent the total loss for the sequence. The results of the top 10 teams with the baseline score are listed in Table 3

In Table 4, we show the dynamic leaderboard progression of tricks and techniques which contributed to our final solution. The analysis was conducted after the competition was finalized and did not include 2% of the corrupted images. Table 3 shows the Private Score on the Hakuna Ma-data challenge where it can be seen that our system improved the baseline score 9x.

We further compare our system against previously published work on the earlier version of the data set. Our system shows significant improvements (see Table 5) reducing the previously reported error rates. In addition, our system performed better than single human annotators, trailing only to human-in-the-loop approach.

## 5. Discussion

**Comparison to previous work**. Previously proposed systems [2, 3] are having two main drawbacks for real life application. First, they use two stage solutions: a) to detect empty images and b) to classify animals. Our contribution is a single step solution. Secondly, previous work was not capable to handle presence of multiple species on the same image as they used softmax output and we have utilized sigmoid function which enables multilabel classification. Table 5 shows that our proposed system surpasses previously reported results by a significant margin by using only a single epoch compared to 50 and 70 in [2] and [3], respectively.

**Augmentation not needed.** Our system features models with low complexity which are fast in terms of both training and inference time. We have used no augmentation except random horizontal flips which allowed us to train for only one epoch. Excluding common augmentation techniques from our pipeline (random crop, shear and rotation) is essential because many objects of interest are not centered and are practically visible (animal entering camera's field of view) which if it is cropped would introduce noise and would require more time to converge.

| Submission evaluation | AggLogLoss |
|---|---|
| Efficient Net B1 | 0.00789 |
| Efficient Net B3 | 0.01102 |
| SE-ResNext-50 s9 | 0.00830 |
| SE-ResNext-50 s10 | 0.00800 |
| SE-ResNext-50 random | 0.00610 |
| SE-ResNext-50 + flip | 0.00592 |
| Ensemble 1 (mean) | 0.00555 |
| Ensemble 2 (gmean) | 0.00582 |
| Ensemble 3 (class aware avrg.) | 0.00550 |

Table 4. Post-competition evaluation on Season 11. The 2% of the images were found corrupted and they are excluded from this analysis.

**Surpassing human performance in embedded edge device.** While comparing our system we significantly improved the current state-of-the-art on Snapshot Serengeti data in all the cases. Additionally, our system performed better than single human annotators on recognizing images without animals, trailing only to human-in-the-loop approach.

| Study | Acc | Empty Acc | Epochs |
|---|---|---|---|
| Willi [3] | 93.4% | 96.0% | 70 |
| Norouzzadeh [2] | 93.8% | 96.8% | 50 |
| Human [4] | 96.6% | 96.6% | - |
| Human-in-the-loop [3] | 99.78% | 99.78% | - |
| **ours** | **94.3%** | **97*%** | **1** |

Table 5. Comparison to other studie. * denotes the result provided by the challenge organizers

## 6. Conclusion

The article describes our winning submission to the "Hakuna Ma-data - Serengeti Wildlife Identification challenge". Proposed system achieved an outstanding score of 0.000531 AggLogLoss while recognizing empty images with 97% Accuracy.

Furthermore we briefly discuss the benefits that biologists might obtain by using such system including but not limited to: 1) reducing manual workload over annotations, thus increasing the efficiency in biology research and 2) minimizing the number of images that have to be processed as there is a lightweight model available which can clean false positives directly on the camera trap device.

All the source code and models weights freely available at https://github.com/drivendataorg/hakuna-madata/tree/master/1st%20Place

## 7. Future Work

In future work we would like to study a problem of poor generalisation to new locations or geographic regions. We expect that our approach should not suffer from generalisation issues, since the models are trained only for single epoch. Additionally we would like to create a system that can handle novel classes. Lastly, deployment of a system with human level accuracy, small power consumption and low complexity directly on the edge device might help to reduce the number of taken images.

## Acknowledgements

This work was supported by the European UnionFLs Horizon 2020 research and innovation program under the Marie Sklodowska-Curie grant agreement No. 642241 to MV. LP was supported by the Ministry of Education, Youth and Sports of the Czech Republic project No. LO1506, and by the grant of the UWB project No. SGS-2019-027.